%% file: RAIL-Principles.tex
\documentclass{article}
\usepackage{arxiv}

\usepackage{graphicx}
\usepackage{wrapfig}
\usepackage{pgfplots}
\usepgfplotslibrary{polar} 
\pgfplotsset{compat=newest}
\usepackage{makecell}
\usepackage[numbers]{natbib}
\usepackage{hyperref}
\hypersetup{
    colorlinks=true,
    citecolor=teal,
    linkcolor=violet,
    urlcolor=teal,
    pdftitle={The RAIL Principles for Neurosymbolic AI: Reasoning, Assurances, Interfacing and Learning},
    pdfauthor={Gnese Chiatti,
Michael Cochez,
Cristina Cornelio,
Sebastijan Dumancic,
Artur d’Avila Garcez,
Luis C. Lamb,
Lia Morra,
Mathias Niepert,
Robert Peharz,
Alberto Speranzon,
Maarten Stol,
Annette Ten Teije,
Thiviyan Thanapalasingam,
Frank Van Harmelen,
Emile Van Krieken,
Antonio Vergari,
Benjie Wang
},
    pdfkeywords={neurosymbolic AI, learning, reasoning},
}

\begin{document}

\title{The RAIL Principles for Neurosymbolic AI: Reasoning, Assurances, Interfacing and Learning}

\author{
{Agnese Chiatti}\\
{Politecnico di Milano}\\
\texttt{\small agnese.chiatti@polimi.it}
\And
{Michael Cochez}\\
\makecell{ELLIS Institute Finland \& \\ \AA{}bo Akademi University}\\
\texttt{\small michael.cochez@abo.fi}
\And
{Cristina Cornelio}\\
{Samsung AI}\\
\texttt{\small c.cornelio@samsung.com}
\And
{Sebastijan Dumancic}\\
{Delft University of Technology}\\
\texttt{\small S.Dumancic@tudelft.nl}
\And
{Artur d'Avila Garcez}\\
{City St. George's, University of London}\\
\texttt{\small a.garcez@city.ac.uk}
\And
{Luis C. Lamb}\\
{Stony Brook University}\\
\texttt{\small luislamb@acm.org}
\And
{Lia Morra}\\
{Politecnico di Torino}\\
\texttt{\small lia.morra@polito.it}
\And
{Mathias Niepert}\\
{University of Stuttgart}\\
\texttt{\small mathias.niepert@ki.uni-stuttgart.de}
\And
{Robert Peharz}\\
{Graz University of Technology}\\
\texttt{\small robert.peharz@tugraz.at}
\And
{Alberto Speranzon}\\
\makecell{Lockheed Martin, \\ Advanced Technology Labs}\\
\texttt{\small alberto.speranzon@lmco.com}
\And
{Maarten Stol}\\
\makecell{BrainCreators \& \\ Vrije Universiteit Amsterdam}\\
\texttt{\small maarten.stol@braincreators.com}
\And
{Annette ten Teije}\\
{Vrije Universiteit Amsterdam}\\
\texttt{\small annette.ten.teije@vu.nl}
\And
{Thiviyan Thanapalasingam}\\
{University of Amsterdam}\\
\texttt{\small t.singam@uva.nl}
\And
{Frank van Harmelen}\\
{Vrije Universiteit Amsterdam}\\
\texttt{\small Frank.van.Harmelen@vu.nl}
\And
{Emile van Krieken}\\
{Vrije Universiteit Amsterdam}\\
\texttt{\small e.van.krieken@vu.nl}
\And
{Antonio Vergari}\\
{University of Edinburgh}\\
\texttt{\small avergari@exseed.ed.ac.uk}
\And
{Benjie Wang}\\
{UCLA}\\
\texttt{\small benjiewang@cs.ucla.edu}
}

\setshorttitle{The RAIL Principles for Neurosymbolic AI}
\setshortauthors{Chiatti et al.}

\maketitle

\begin{abstract}
Neurosymbolic AI systems that integrate machine learning and symbolic reasoning are rapidly gaining attention. They complement the data-intensive statistical approaches of neural networks and language models with symbolic reasoning algorithms to function in high-stakes domains or in low-data regimes that characterize many real-world applications. We argue that the neurosymbolic combination of machine learning and formal reasoning is not a niche approach within AI, but rather includes many already successful techniques that are of crucial importance to the development of reliable, efficient and, ultimately, trustworthy systems. This perspective prompts a re-examination of the design of current AI systems. We show that many leading AI systems, 
{including some that are not traditionally  considered as neurosymbolic,} 
can be analysed from the perspective of four principles of neurosymbolic AI design: \emph{Reasoning}, \emph{Assurances}, \emph{Interfacing} and \emph{Learning} (RAIL). Applying the RAIL framework offers a unified view of seemingly disparate AI systems, ranging from physics-aware machine learning to {neuro-guided search (such as Google DeepMind's Alpha-* suite)}, causal learning and tool-augmented Large Language Models. Importantly, the  RAIL principles 
will enable engineers to make better-informed and more principled decisions about the design and deployment of production-level AI systems. 
In this article, we introduce the RAIL principles, examine how they can be applied across major areas of AI, and illustrate how they may guide practitioners to integrate neurosymbolic methods into next-generation AI technologies.
\end{abstract}

\section{Neurosymbolic AI} 
\label{sec:intro}

Neurosymbolic AI brings together the statistical nature of machine learning and the  rigorous logical nature of reasoning \cite{GarcezL23}. The field has grown in relevance as it became clear that the combination of learning and reasoning within a formal semantic framework is crucial for AI to reach its full promise in science and engineering \cite{besold2022neural,GLG2009neural}.

Industrial AI systems that ingest sensor, text, and video streams are typically assembled from ad‑hoc, task‑specific modules that are designed and optimized independently. This divide‑and‑\allowbreak conquer approach accelerates development but results in non‑compositional pipelines with no formal guarantees on overall behavior, leading to error propagation, discontinuities at module interfaces, and difficulties in interpreting the reasoning of generative models. The persistent problem of LLM hallucinations, four years after the release of ChatGPT, underscores the need for a more principled solution. Neurosymbolic AI addresses this gap by embedding sound symbolic reasoning into machine learning components, enabling verifiable constraints on model outputs and allowing the system to reason about what has been learned. We therefore frame neurosymbolic AI around four core principles, Reasoning, Assurances, Interfacing, and Learning (RAIL), which together provide a unified blueprint for building reliable, explainable, and compositional industrial AI pipelines. 
{We will show that the RAIL principles also apply to systems that may not traditionally be considered neurosymbolic, but which do inject knowledge into their learning architecture, such as physics informed neural networks and industrial modular pipelines.}

\begin{figure}[b]
    \centering
     \includegraphics[width=0.9\linewidth]{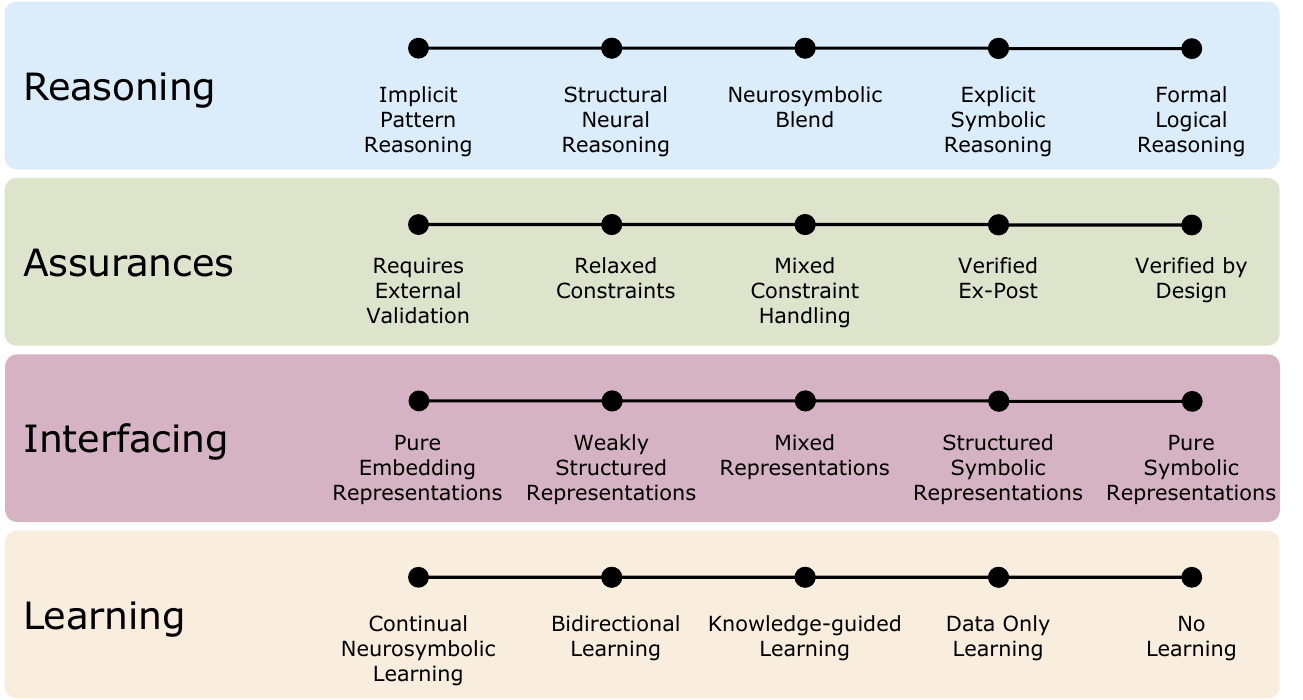}
    \caption{The four \textsc{RAIL} principles of Neurosymbolic AI. Each principle exists on a qualitative spectrum, with intermediate positions showing different combinations of neural and symbolic approaches.} 
    \label{fig:rail-dimensions}
\end{figure}

\section{The RAIL Principles of Neurosymbolic AI}
\label{sec:rail}

\subsection{Reasoning } 
\label{sec:reasoning}

Reasoning, i.e., chaining a series of inference steps to derive a conclusion, has been  essential  in the study of machine intelligence. It has been deployed successfully in AI systems for  question-answering, decision support, planning, and story understanding, among others. Reasoning can be deductive (\textit{Which consequences follow from a premise?}), abductive (\textit{Which premise best explains an observation?}), inductive (\textit{Which conclusion is justified by given data?}) or analogical (\textit{What are the correspondences between sequences of inference steps?}). Such reasoning capabilities have typically been realized through explicit symbol manipulation in a wide variety of formal frameworks \cite{van2008handbook}. 

The benefits of sound reasoning are a key advantage of neurosymbolic AI \cite{GLG2009neural}, 
based on results in classical and non-classical logics. 
Through representing modal, temporal, and probabilistic reasoning, neurosymbolic approaches support the representation of knowledge, actions, and beliefs of interacting agents \cite{besold2022neural,GLG2009neural}. 
Beyond logic, the symbolic component can, in fact, be any computer program, e.g. used as a tool by an LLM for the purpose of reasoning. 
Recent advances in LLMs show how GenAI displays an informal, emergent, and latent form of reasoning {based on a prompting process known as \textit{chain of thought}}. {Our design of the Reasoning dimension of the RAIL principles (see Fig. \ref{fig:rail-dimensions}) acknowledges this possibility, }
with  such entirely implicit reasoning on one end---e.g. emergent reasoning by similarity---and \emph{formal reasoning} on the other. In between,  neurosymbolic systems such as DeepProbLog~\cite{DeepProbLog} and Logic Tensor Networks \cite{LTN}  explore a wide scale of implicit and explicit reasoning. This includes neurosymbolic systems with \emph{explicit symbolic reasoning}, where neural models call symbolic computation (e.g. generating code in Python and calling an interpreter), and \textit{structural neural reasoning} \cite{GLG2009neural} where reasoning is encoded in the structure and weights of a neural network, with a variety of neurosymbolic blends in between \cite{besold2022neural}.  
{Our Reasoning dimension also {includes} systems beyond those typically considered as neurosymbolic, {specifically we consider systems that} encode background knowledge, either explicitly in symbols or implicitly in their architecture.}
{Despite rapid advances in LLM reasoning, recent results \cite{barman2026pricemeaningsemanticmemory} show that the geometric structure that enables semantic generalisation also makes interference, forgetting, and false recall {inevitable}. These inherent limitations of systems on the left side {of} the Reasoning spectrum imply that future systems will have to inhabit the entire spectrum for fully reliable inference.}

\subsection{Assurances} 
\label{sec:assurances}
This dimension of RAIL represents the  \textit{assurances} that a neurosymbolic system can provide with respect to given \textit{requirements}. These can take the form of a formal specification that designers and domain experts elicit for safety-critical systems, formalized using one of the many suitable logics (propositional, temporal, or spatio-temporal logic), or can be unit-tests in the case of programs. The specifications may formalise diverse non-functional requirements too: safety (``the system must never enter an unsafe state''), performance (``the system must  reach a desired state in a given time''), or reliability (``given a set of initial conditions, the system must always reach a target set of states''). 

For systems that learn, particularly those using deep neural networks, such requirements are notoriously difficult to check. {While the field of neural network verification has devised several techniques to ensure that certain required properties like robustness and bounded behaviour are met, verification guarantees in these systems often rely on simplified assumptions or computationally expensive procedures.}
On one side of the Assurances spectrum 
we find neural components trained without any requirements, 
thus requiring external validation. On the other side of the spectrum, we find neural components that are \textit{verified by design}, i.e. the requirements are baked in the network architecture. In the middle 
are the approaches that treat requirements as  soft or hard symbolic constraints during learning. This is typically done by relaxing the logical representation of a requirement into differentiable form, e.g. using fuzzy logic \cite{van2022analyzing, LTN}, making it suitable for training, runtime monitoring and inference-time reasoning. It also includes neural networks with \textit{shield layers} that assign zero probability to the output configurations that may violate a given constraint, e.g. algorithms to steer LLMs to guarantee that they produce outputs satisfying a given grammar or syntactical structure \cite{ctrl-g}. 
Finally, neural networks' properties or outcomes may be verified ex-post, after the system has been trained. 
{Inherent limitations on assurances by neural systems shown in \cite{barman2026pricemeaningsemanticmemory} imply that symbolic approaches to assurances will remain essential.}
{Overall, balancing these assurances against their inherent limitations and scalability barriers remains a core open challenge for neurosymbolic architectures especially in complex real-world deployments.}

\subsection{Interfacing } 
\label{sec:interfacing}

Neural networks have been criticized for not being \textit{interpretable} or \textit{explainable},  understood as the capacity for internal processes or network outcomes to be inspected by or explained to humans. These requirements are particularly important when decisions made by AI are high-risk or when outcomes may induce discrimination. Interpretability and explainability presuppose a \textit{communication channel} from the system to the user.
Conversely, robust and generalizable reasoning often depends on the ability to communicate \textit{domain knowledge} to the system in a well-defined form, requiring a channel also from the user to the system. These bidirectional requirements motivate the notion of \textit{interfacing} as a RAIL principle: establishing shared representations that support meaningful information exchange between agents, whether human or artificial. 

Like the other RAIL principles, interfacing is characterised on a spectrum  (Fig. \ref{fig:rail-dimensions}). At one end are the latent representations in a neural network, which excel at capturing high-dimensional statistical structure but offer limited direct inspection or user intervention. Moving along the spectrum, weakly structured representations introduce partial organization, e.g., attention patterns or neuron groupings, without committing to a defined semantics. Mixed representations, including graphs and  causal models,  support intervention while remaining compatible with subsymbolic systems.  
Similarly, natural languages have syntax and semantics, but are also used to convey other information (e.g., intentions, emotions) that often reside on a subsymbolic level. 
Further along, structured symbolic representations such as logic programs, annotated code, or structured data formats 
prioritize explicit compositional structure. Finally, pure symbolic representations, grounded in formal knowledge representation with unambiguous semantics, can be used to make interfacing requirements explicit, albeit at the cost of often requiring representations to be manually-specified.
Each position on this spectrum affords distinct trade-offs in abstraction, expressivity, and handling of uncertainty in the interfaces from user to system (guidance) and from system to user (explanation). 

\subsection{Learning} 
\label{sec:learning}

The \textit{learning} dimension of RAIL captures the degree to which a system applies machine learning 
to shape its behaviour. 
At one end of the spectrum, symbolic systems use pre-defined knowledge \textit{without learning}, and depend on knowledge engineering rather than statistical generalisation from examples. Deep learning models acquire their behaviour through \textit{data-only learning} by 
extracting statistical regularities, which often requires extensive data and offers limited control over what is learned. 
Neurosymbolic systems go beyond learning from data alone and combine the use of structured knowledge to guide learning and decision-making~\cite{Rueden2023informed}. 
Many systems employ some form of \textit{knowledge-guided learning} to shape learned representations, improve generalisation, or reduce data requirements \cite{Rueden2023informed}.  Knowledge can be added implicitly via architectural inductive bias---Convolutional Neural Networks make implicit use of knowledge about invariance---or by curating and augmenting data with informed transformations.  
Neurosymbolic systems enable a more explicit use of knowledge, that may include encyclopedic facts, taxonomies, abstract rules, logical constraints, program sketches or  expert scientific structures like 
differential equations in physics-informed machine learning~\cite{raissi2019physics}.  
Incorporating knowledge can bring substantial improvements in data efficiency compared to purely neural baselines~\cite{LTN,DeepProbLog,cory2024evolving}.  Further towards the end of the spectrum, some systems  learn to reason, that is to learn the very mechanisms that operate over logical formalisms rather than treating logic as an external interpreter. 
Such systems do not treat knowledge as fixed  background knowledge, but learn to evolve it as part of what we call \textit{bidirectional learning}~\cite{NIPS2003_34766559}. 
Finally, models may be fine-tuned after training through feedback loops such as reinforcement learning or self-training. In \textit{continual neurosymbolic learning},  learning takes place over time with evolving knowledge to inform the learning process.

\section{Putting Existing AI Systems on RAILs}
\label{sec:areas}

We illustrate RAIL’s usefulness by positioning prominent AI systems within the RAIL spectrum across five key AI domains, discussing the opportunities afforded by moving along the RAIL scales, and summarizing industrial neurosymbolic applications from the perspective of the RAIL principles.

\subsection{Knowledge Discovery} 
\label{sec:knowledge-completion}

Knowledge graphs (KGs) are a standard data model for representing entities and their relationships, governed by an ontology of types and constraints \cite{CSUR-KG-paper}.~Originally designed for data integration and query answering, KGs are increasingly exploited to predict new knowledge, i.e., novel edges between entities, through neurosymbolic techniques.
Two main families of methods address this link prediction task. Embedding‑based approaches {\cite{KGcompletion}} map graph nodes to high‑dimensional vectors, training a loss function that aligns symbolic patterns with geometric ones so that vector similarities can suggest new relations (e.g., discovering chemical bonds). Rule‑based systems such as AnyBURL \cite{AnyBURL} learn explicit rules from the graph and use symbolic reasoning to infer new rules (e.g., drug‑gene‑symptom patterns for drug repurposing). 

\noindent\textbf{Reasoning.}
An embedding-based technique reasons implicitly on the patterns in  vector space (e.g. vector similarity or distances). A drastic shift along the Reasoning dimension is taken by rule-learning systems. Instead of reasoning over geometric patterns in vector spaces, they perform explicit symbolic reasoning applying the rules that were discovered. 

\noindent\textbf{Assurances.}
Only soft assurances with \textit{relaxed constraints} are given on the correctness of newly predicted relations. Minimization of the loss function increases the likelihood of a predicted relation but it does not provide guarantees. Assurances can move from relaxed constraints to verified constraints via ex-post verification of ontological demands, resulting in the removal of predicted relations that violate constraints such as types, cardinality, anti-symmetry. 

\noindent\textbf{Interfacing.}
The combination of a symbolic graph with numerical vector spaces offers a mixed level of  interfacing. In the rule-learning methods, all representations (the graph, the rules and the newly discovered relations) are symbolically represented and expected to be humanly interpretable. By contrast, embeddings are weakly structured with difficult interpretability.

\noindent\textbf{Learning.} When a KG is used simply as a source of data, learning will be data-only learning because no further knowledge is injected. The Learning dimension shifts to knowledge-guided learning when using more informed (domain specific) sampling strategies to learn embeddings, e.g. type constraints \cite{VISE} or other ontological properties of the graph \cite{box-embeddings}. Rule-based learning is typically bi-directional. They first learn the rules from the graph data and then apply the rules to the graph data to discover new relations.

\setlength{\columnsep}{2pt} 
\begin{wrapfigure}[8]{r}{0.3\textwidth} 
\centering
\vspace{-2em}
\resizebox{0.3\textwidth}{!}{\input{radarplot-knowledge-discovery.tex}}
\label{fig:radarplot-knowledge-discovery}
\end{wrapfigure}

\noindent\textbf{Lessons for RAIL.}
The two families of knowledge discovery methods are located very differently on the RAIL dimensions. For both families, assurances would move from relaxed constraints to verified constraints if verification techniques were applied to ontological knowledge. 
{The {radar plot} shows that embedding methods (blue) place more emphasis on learning, while the symbolic rule-learning methods (red) provide more assurances and {better} interfacing.}

\vspace{2em}

\subsection{Neuro-Guided Search}
\label{sec:structured-problem-solving}

Neuro‑guided search is a neurosymbolic approach where neural networks modify a structured symbolic representation, often using reasoning to ensure correctness. As a case-in-point, the Google DeepMind Alpha family of systems (AlphaGo, AlphaFold, AlphaProof, etc.) combine neural learning with Monte~Carlo Tree Search{~\cite{silver2016mastering}}. In these systems, the neural component actively navigates and expands the symbolic problem structure (game trees, proof trees) rather than invoking external tools. This tight integration of neural and symbolic parts is the core inventive step behind their state‑of‑the‑art performance across domains.

\noindent\textbf{Reasoning.}
Neuro-guided search systems employ explicit structured representations of the problems they solve: AlphaGo explicitly represents the game state, neuro-guided program synthesis methods represent programs as symbolic structures, AlphaProof and AlphaGeometry explicitly represent the proofs in the LEAN  programming language. 
Consequently, they all support explicit formal reasoning about their outputs. The outcome of the reasoning step can be explicitly used during the learning process.
At the same time, the learned problem-solving process itself, e.g. the action selection in AlphaGo or the lemma proposition in AlphaProof, is implicit and opaque. 

\noindent\textbf{Assurances.}
By using structured representations, these systems offer strong assurances by design.
For instance, AlphaProof guarantees proof correctness through LEAN. The other Alpha family systems are prevented from exploring invalid programs or strategies that violate the  game rules. 

\noindent\textbf{Interfacing.}
Neuro-guided search systems have high interfacing from user to system: the user can provide new lemmas to AlphaProof and AlphaGeometry or new game rules to AlphaGo.
However, the interfacing from system to user is limited because the decisions of the system take place within the black box neural network. AlphaProof cannot explain why it proposes a specific lemma to be considered, and AlphaGo cannot explain why a particular move is promising.

\noindent\textbf{Learning.}
Learning in most neuro-guided search systems is  knowledge-guided: the neural components learn how to make decisions constrained by the structured representations. For instance, in AlphaGo, the neural network learns to propose actions, and the rules of the game constrain which actions are possible; neuro-guided program synthesisers propose completions for abstract-syntax trees constrained by the syntax of a programming language. The structured problem representation remains unchanged while all the learning happens in the neural network. In some systems, however, learning is bidirectional. For instance, AlphaProof changes the representation of the problem when proposing new lemmas. That new representation can also be used in the solution of other problems. AlphaTensor similarly explicitly changes the algorithm structures through decompositions ~\cite{fawzi2022discovering}. 

\noindent\textbf{Lessons for RAIL.}
The Alpha systems highlight key trade-offs across the RAIL dimensions that inform the design of neurosymbolic systems. (1)~\textit{Reasoning--Assurance trade-off:} The progression from AlphaGo to AlphaProof illustrates how stronger formal assurances necessitate more explicit reasoning components. AlphaGo's game rules provide implicit constraints, whereas AlphaProof's integration with LEAN enables machine-verified proofs, showing that domains requiring high-stake guarantees benefit from tighter coupling with formal verification systems. (2)~\textit{Interfacing asymmetry:} A consistent pattern emerges wherein interfacing \textit{into} these systems (modifying rules, providing lemmas) is high, while interfacing \textit{out of} the systems (explaining decisions) remains limited. The outputs are transparent and verifiable, but decision-making is opaque. (3)~\textit{Learning at the Knowledge-Guided level:} in most Alpha systems, learning occurs within the neural component and the symbolic structure remains fixed. In Alpha Proof and AlphaTensor's bidirectional learning, however, the symbolic representation evolves with new lemmas and decompositions, indicating that richer neurosymbolic integration is achievable when the symbolic element is flexible.  
        
\subsection{Tool Use by Language Models} 
\label{sec:llm-tools}

A  rapidly evolving area where neurosymbolic principles have re-emerged is the coupling of LLMs with computer program, either through \textit{program synthesis}, where LLMs generate executable code from partial specifications in natural language or examples, or \textit{via tool usage}, where LLMs invoke existing APIs or database queries. Program synthesis originated in symbolic AI, emphasizing explicit representations and guaranteed search \cite{cropper2022inductive}. Although purely neural LLM methods \cite{austin2021program} have emerged, current systems are increasingly neurosymbolic, combining LLMs with programs, tests,
grammars, interpreters, and verification engines.  Tool‑augmented frameworks like ReAct \cite{yao2022react}  exemplify this shift, operating where the model reasons about the problem, selects a concrete action, executes it within an external tool, and refines its behavior based on the result. {This coupling is valuable because part of the LLM tasks are delegated to components with explicit semantics. However, reliability depends not only on tool correctness, but also on which tool the model selects and on how it integrates results from different calls, }{both of which are often brittle.}

\noindent\textbf{Reasoning.} LLMs that use tools go beyond implicit pattern completion. While the base language model still performs latent statistical inference, the explicit generation and execution of programs or structured actions introduces symbolic semantics into the reasoning process. In the case of ReAct, reasoning is made more explicit by alternating between deliberation steps and symbolic actions (tool calls and environment interactions). {Interleaving reasoning and actions can reduce hallucination and error propagation, but in longer {sequences} early tool-selection errors or misinterpreted observations may still cascade into later steps.} 

\noindent\textbf{Assurances.} Tool use enables guarantees currently unavailable to LLMs. At one end, execution, compilation, and unit testing provide the possibility of empirical checks to filter invalid outputs. More advanced approaches integrate formal verification {\cite{Cai2025}}, differential testing, or oracle-guided refinement, yielding assurances about functional correctness.  
{In these approaches, selection and integration of external tools is crucial but remains difficult across the latent/explicit barrier.}

\noindent\textbf{Interfacing.} LLMs with tools offer more interfacing possibilities than standalone LLMs. 
Programs, grammars and API calls provide compositional, inspectable and executable interfaces between neural and symbolic agents, as well as human users. 
Visual programming systems like ViperGPT \cite{suris2023vipergpt} extend to multimodal settings where symbolic programs orchestrate learned perception modules from a query.
{However, interfaces that integrate LLMs with tools can introduce a degree of brittleness, as APIs, schemas, and grammars often require exact syntax.}

\noindent\textbf{Learning.} Most current tool-augmented systems combine pre-trained foundation models with prompting, in-context learning, or reinforcement learning from execution feedback. Toolformer-style approaches{\cite{schick2023toolformer}} explicitly learn when and how to invoke tools. Other AI agents will tackle a task by generating a function from a few examples, testing it against a differentiable oracle that provides feedback for further refinement, e.g. agents tackling the \textit{Abstraction and Reasoning Corpus} (ARC) challenge \cite{wei2025codearc}. These bidirectional learning cycles closely mirror neurosymbolic learning paradigms where the outcomes of symbolic execution guide neural network updates, and the neural networks' heuristics guide symbolic search.

\setlength{\columnsep}{2pt}  
\begin{wrapfigure}[11]{r}{0.3\textwidth} 
\centering
\vspace{-3em}
\resizebox{0.3\textwidth}{!}{\input{radarplot-LLMs-tools}}
\label{fig:radarplot-LLMs-tools}
\end{wrapfigure}

\noindent\textbf{Lessons for RAIL.} Tool-augmented LLMs offer design directions for broader advances in neurosymbolic AI. The integration of structured programs and representations with LLMs achieves a trade-off between bottom-up learning from data and top-down enforcement of constraints. Tools can also serve as a method to check for correctness of LLM-based systems. Combining natural language with other data modalities impacts the interfacing dimension of RAIL, increasing interoperability among tools that require heterogeneous input and output representations.  
{The {radar plot} shows the differences between LLMs with tools (red) and without (blue).} Tool-augmented LLMs expose scalability barriers for neurosymbolic AI. Tool-selection instability, cascading reasoning errors, and brittle symbolic interfaces limit reliability, especially as tools grow in number and diversity. These limitations suggest that scaling and reliability depend on the right balance between symbolic components for explicit reasoning and controllability, and subsymbolic components for flexibility and generalisation \cite{zhang2025position}. 

\subsection{Causal Learning and Reasoning} 
\label{sec:causal}

Causal modeling \cite{Pearl2000} formalizes the representation and learning of causal knowledge, most often using causal graphs in which nodes denote variables and edges denote causal relationships. Within this framework, two central tasks are: \textit{causal discovery}, which seeks to recover the graph structure from data and \textit{causal inference}, which estimates the effects of interventions by combining data with partial knowledge of the graph. Because causal methods integrate statistical learning with explicit structural constraints and domain knowledge, they naturally embody neurosymbolic principles.  
\textit{Identifiability} specifies when causal effects can be uniquely determined from observed data and graph assumptions, so progress in this area has depended on fusing symbolic representations and formal reasoning with statistical estimation rather than relying solely on black‑box models.

\noindent\textbf{Reasoning.} Causal reasoning is formalized through a three-level causal hierarchy: 
association, intervention, and counterfactuals \cite{Pearl2018}.  
\emph{Association} concerns reasoning about statistical dependencies and predictive tasks in standard machine learning.
Various limitations of machine learning, such as poor extrapolation, and shortcuts, have been studied from a causal perspective.
A causal perspective at the level of \emph{Intervention} opens the possibility of systems with improved generalization and extrapolation capabilities, in particular under shifts in data distribution. One of the 
most widespread applications of intervention analysis is in econometrics and e-commerce~\cite{sharma2015estimating}.
Understanding and designing biological pathways 
is another prominent example. 
The third level of Pearl's causal hierarchy is counterfactual reasoning, seeking to answer the question \textit{If X had not occurred, would Y have been the case?}, a question closely connected with  
explainability in neural networks \cite{White2023}.

\noindent\textbf{Assurances.} 
Probabilistic inference under observational, interventional and counterfactual distributions guarantees that, as long as the Structural Causal Model (SCM) specification and probabilistic inference routines are correct, strong assurances can be provided. 
Computational complexity issues aside, probabilistic inference also provides an opportunity to tackle epistemic challenges with assurances. The \emph{do-calculus} \cite{Pearl2000} is a complete and consistent algorithm to decide whether interventional queries can be answered from the observational distribution and causal structure alone. 

\noindent\textbf{Interfacing.} The intuitive graphical structure of causal models makes them easy to interface with on a qualitative level. Furthermore, causal models are naturally modular, in that it is not only possible to, but natural to modify the causal structure of subgraphs. 

\noindent\textbf{Learning.} Causal learning often involves discovering the causal structure (graph), or causal mechanisms and probability distributions from data. A key constraint in causal learning comes from fundamental \emph{identifiability} barriers, which restrict what can be learned from available data. This motivates the use of knowledge guided learning approaches. Another important challenge is to connect causal reasoning, which is typically defined over a set of interpretable variables, with high-dimensional unstructured data such as images. This is the goal of \emph{causal representation learning}, which aims to mine causal representations from unstructured data \citep{scholkopf2021toward}.

\noindent\textbf{Lessons for RAIL.}
Causality exemplifies a neurosymbolic design pattern in which a symbolic structure, typically a causal graph, mediates between data-driven learning and formal reasoning. On {Reasoning}, causality explicitly distinguishes associative prediction from intervention and counterfactual analysis. For {Assurances}, it provides guarantees when causal queries are answerable from data and assumptions, and when they are fundamentally underdetermined. On {Interfacing}, causal models offer modular representations that support human intervention. 
Finally, on  {Learning}, causal discovery and representation learning illustrate knowledge-guided and bidirectional learning where statistical estimation is constrained by symbolic structure and vice versa.

\subsection{Physics-Aware Machine Learning } 
\label{sec:pinns}

Machine learning for science and engineering has developed largely in parallel with neurosymbolic AI, yet the overlap is substantial when viewed through the lens of the \textsc{RAIL} principles. Neurosymbolic AI traditionally injects \emph{explicit structure} via logical rules, programs, or constraints. Physics-aware ML injects structure via \emph{physical laws}~\cite{cory2024evolving}, \textit{symmetries}~\cite{akhound2023lie}, \textit{conservation principles}~\cite{raissi2019physics}, and \textit{mathematical operators}~\cite{batzner2022equivariant}. 
These ingredients are not always described as symbolic, but play the same conceptual role. They encode human knowledge in declarative form, constrain what solutions are admissible, and support more reliable generalization beyond the training distribution.

\noindent\textbf{Reasoning.}
Many \textit{AI for Science} methods employ forms of reasoning that closely mirror neurosymbolic inference.
Forward simulation is akin to \emph{deduction}: given assumptions (initial conditions, parameters), derive consequences (system trajectories).
Inverse problems and parameter identification resemble \emph{abduction}: infer latent causes or mechanisms that explain observations.
Changing boundary conditions or forcing terms produce \emph{counterfactuals}, analogous to interventions in causal reasoning.
Multi-physics models combine multiple modules (e.g. fluid and structure), similar to compositional reasoning, where complex behaviour is assembled from reusable parts.

\noindent\textbf{Assurances.}
Assurances may come from physical structure and invariances.
Some assurances can be enforced \emph{by design} through architecture \cite{Giunchiglia2024CCN+} rather than through a loss function.
For instance, equivariant neural networks~\cite{batzner2022equivariant} 
guarantee
the encoding of symmetries (e.g. rotation, translation).
Hamiltonian architectures~\cite{greydanus2019hamiltonian}  enforce conservation laws by construction.
As in neurosymbolic systems, these assurances rely on the correctness of the assumed structure (e.g. the conserved quantity) and can degrade under misspecification, discretization artifacts, or missing physics.

\noindent\textbf{Interfacing.}
Neurosymbolic AI emphasizes interfaces that let experts state \emph{what} they know without prescribing \emph{how} to compute it.
Science has similarly rich declarative languages: differential equations, symmetries, conservation laws, and energy functions.
These are natural symbolic interfaces between humans and models. Thus, work in neurosymbolic AI on intermediate representations (constraints, knowledge graphs) suggests a direction for scientific ML: enable incremental, partially specified, and revisable scientific knowledge to guide learning and reasoning.

\noindent\textbf{Learning.}
In neurosymbolic learning,  knowledge is incorporated either in the network architecture \cite{Giunchiglia2024CCN+} or by augmenting the training objective with penalties for violating symbolic constraints. 
Physics-aware ML follows this pattern, with physical models as symbolic knowledge. In physics-informed approaches~\cite{cuomo2022scientific}, learning includes penalties for inconsistency with physics equations, boundary conditions, or conservation laws. 
Such physics-based terms encourage solutions that respect known physical structure, even when observations are sparse or noisy. In \cite{cuomo2022scientific}, knowledge-based penalties are implemented as soft constraints using weighted regularization terms. 

\noindent\textbf{Lessons for RAIL.}
Physics-aware ML is in many ways neurosymbolic, as it integrates learned models with explicit, human-authored structures.
Viewing it through the \textsc{RAIL} lens helps clarify design trade-offs: where to enforce knowledge (loss function or network architecture), how to represent such constraints (rules or operators), and how to learn such constraints (PDEs or logic). 

\subsection{Industrial Neurosymbolic Architectures} 
\label{sec:industrial}

The neurosymbolic paradigm in industrial AI has an additional driver: architectural necessity. Mature software products typically do not deploy a single neural model
but consist of a composite software stack in which pragmatic demands on control and flexibility,
together with the economy of product maintenance, drive the designs toward modular architectures. These architectural patterns are implicitly neurosymbolic: deep learning components 
that rely on learning from data combine with surrounding procedural glue code addressing logical requirements and encapsulating domain knowledge. 
Industry has matured this design pattern over the last decade largely independent of 
academic neurosymbolic AI. These  software stacks function effectively as they allow ML engineers to enforce strict symbolic controls on stochastic neural components. Consequently, the lack of formal neurosymbolic methods is not a bottleneck that prevents industry from deploying products, using pragmatic, albeit disjointed, integration.
The value proposition of formal neurosymbolic research for industry lies in providing more principled integration.
The joint academic-industrial roadmap therefore concerns the transformation of these ad-hoc architectures into formal design patterns~\cite{vanBekkum}. 
The goal is to replace brittle glue code and symbolic knowledge interfacing in the stack with the expressive, differentiable methods highlighted in previous sections
(e.g., replacing a pre-processing script with a symbolic regularizer, or a post-processing heuristic with a differentiable reasoner). 
This upgrade path would move the system from an assembly of black boxes towards differentiable interfacing, ideally without sacrificing the engineering benefits of modularity.
\\
\noindent\textbf{Lessons for RAIL.} 
RAIL principles are clearly present in industrial AI architectures. Learning modules are data only, or knowledge-guided by domain informed regularization. Interfacing ranges from using  embeddings to structured symbolic representations, and reasoning in the form of symbolic glue code is typically informal. 
Assurances remain the most challenging RAIL dimension for such architectures. 
But this is also why assurances based on increasingly formal yet differentiable methods are how the neurosymbolic paradigm will have the biggest impact on industrial AI. 

\section{Conclusion: The Neurosymbolic Path Forward} 
\label{sec:path-forward}
We have shown that many high-impact AI systems, such as those from the Alpha* family,  can be re-interpreted as neurosymbolic,  unwittingly following the RAIL principles.
The RAIL principles 
map the trade-offs and synergies that shape neurosymbolic architectures. They define a four‑dimensional design space in which the inter-dependent axes constrain viable neurosymbolic architectures. Mapping existing systems onto this space revealed such trade‑offs and synergies.
{By visualising these relationships,} RAIL offers a  framework for exploring new design opportunities, 
guiding future AI research and industrial deployments. 
{We have provided only a qualitative treatment of the RAIL dimensions. Turning them into a practical tool will require operational criteria, 
formal definitions, and concrete metrics. We {trust} the community will rise to this challenge.}

\section*{Acknowledgements}
This paper is the result of Dagstuhl Seminar 25452: A Roadmap Towards Practical Applications of Neurosymbolic Learning and Reasoning.
We thank the other participants for valuable discussions: Byron Cook (Amazon Web Services \& University College London),  Egor Kostylev (U. of Oslo), Robin Manhaeve (KU Leuven), Guy Van den Broeck (UCLA), Maria-Esther Vidal (Leibniz U. Hannover), and Ute Schmid (U. of Bamberg). 
{We thank the anonymous reviewers for valuable comments.}  

\bibliographystyle{plainnat}
\bibliography{software,sample-base}

\end{document}

%% file: radarplot-knowledge-discovery.tex
\begin{tikzpicture}
\begin{polaraxis}[
    ymax=5,
    xtick={0, 90, 180, 270},
    xticklabel style={font=\Huge},
    xticklabels={R, A, I, L},
    xticklabel style={inner sep=10pt},
    axis line style={gray},
    yticklabel style={text=black!70, font=\Large},
    major grid style={gray},
    minor grid style={gray!40},
    axis on top,
    grid=both,
    minor tick num=1
]

\addplot[color=blue, thick, mark=*, fill=blue!20, opacity=0.5, mark options={fill=blue!80, draw=blue!80, solid}] coordinates {
    (0, 1)   
    (90, 2)  
    (180, 2) 
    (270, 5) 
    (0, 1)
};
\addplot[color=red, thick, mark=*, fill=red!20, opacity=0.5, mark options={fill=red!80, draw=red!80, solid}] coordinates {
    (0, 4)   
    (90, 4)  
    (180, 5) 
    (270, 2) 
    (0, 4)
};

\end{polaraxis}
\end{tikzpicture}

%% file: radarplot-LLMs-tools.tex
\begin{tikzpicture}
\begin{polaraxis}[
    ymax=5,
    xtick={0, 90, 180, 270},
    xticklabel style={font=\Huge},
    xticklabels={R, A, I, L},
    xticklabel style={inner sep=10pt},
    axis line style={gray},
    yticklabel style={text=black!70, font=\Large},
    major grid style={gray},
    minor grid style={gray!40},
    axis on top,
    grid=both,
    minor tick num=1
]

\addplot[color=blue, thick, mark=*, fill=blue!20, opacity=0.5, mark options={fill=blue!80, draw=blue!80, solid}] coordinates {

    (0, 2)   
    (90, 1)  
    (180, 1) 
    (270, 4) 
    (0, 2)
};
\addplot[color=red, thick, mark=*, fill=red!20, opacity=0.5, mark options={fill=red!80, draw=red!80, solid}] coordinates {
    (0, 5)   
    (90, 5)  
    (180, 3) 
    (270, 4) 
    (0, 5)
};
\end{polaraxis}
\end{tikzpicture}